\documentclass[journal]{IEEEtran}

\usepackage{amsmath,amssymb,amstext} 
\usepackage[detect-all=true]{siunitx}  
\usepackage{bbm}  
\usepackage{textcomp}  
\usepackage{bbold}  
\usepackage[pdftex]{graphicx}
\usepackage{caption}
\usepackage{multirow}
\usepackage{array}
\usepackage{float}  
\usepackage{url}
\usepackage{xspace}  
\usepackage{enumitem}  
\usepackage{authblk}  
\usepackage{rotating}
\usepackage{pdflscape}
\usepackage{gensymb}  
\usepackage{todonotes}  
\newcommand{\added}[1]{#1}

\renewcommand{\vec}[1]{\mathbf{#1}}

\newcommand{\ve}{$\dot{\text{V}}_\text{E}$}
\newcommand{\vo}{$\dot{\text{V}}\text{O}_2$}
\newcommand{\vopeak}{$\dot{\text{V}}\text{O}_{2\text{peak}}$}

\title{Temporal convolutional networks predict dynamic oxygen uptake response from wearable sensors across exercise intensities}

\author{Robert Amelard,
        Eric T Hedge, Richard L Hughson
\thanks{R.~Amelard and E.~T.~Hedge contributed equally. (\textit{Corresponding author: Robert Amelard})}
\thanks{R. Amelard is with the KITE-Toronto Rehabilitation Institute, University Health Network, Toronto, Canada and Schlegel-UW Research Institute for Aging, Waterloo, Canada (e-mail: robert.amelard@uhn.ca)}
\thanks{E.~T. Hedge and R.~L. Hughson are with the Schlegel-UW Research Institute for Aging, Waterloo, Canada and University of Waterloo, Waterloo, Canada. R.L.H. is the Schlegel Research Chair in Vascular Aging and Brain Health.}}

\date{}

\begin{document}

\maketitle

\bstctlcite{IEEEexample:BSTcontrol}

\begin{abstract}
Oxygen consumption (\vo) provides established clinical and physiological indicators of cardiorespiratory function and exercise capacity.
However, \vo~monitoring is largely limited to specialized laboratory settings, making its widespread monitoring elusive.
Here, we investigate temporal prediction of \vo~from wearable sensors during cycle ergometer exercise using a temporal convolutional network (TCN). Cardiorespiratory signals were acquired from a smart shirt with integrated textile sensors alongside ground-truth \vo~from a metabolic system on twenty-two young healthy adults. Participants performed one ramp-incremental and three pseudorandom binary sequence exercise protocols to assess a range of \vo~dynamics. A TCN model was developed using causal convolutions across an effective history length to model the time-dependent nature of \vo. Optimal history length was determined through minimum validation loss across hyperparameter values.
The best performing model encoded 218~s history length (TCN-VO2~A), with 187~s, 97~s, and 76~s yielding less than 3\% deviation from the optimal validation loss.
TCN-VO2~A showed strong prediction accuracy (mean, 95\% CI) across all exercise intensities (\textminus22~\si{ml.min^{-1}}, [\textminus262, 218]), spanning transitions from low-moderate (\textminus23~\si{ml.min^{-1}}, [\textminus250, 204]), low-\added{high} (14~\si{ml.min^{-1}}, [\textminus252, 280]), ventilatory threshold-\added{high} (\textminus49~\si{ml.min^{-1}}, [\textminus274, 176]), and maximal (\textminus32~\si{ml.min^{-1}}, [\textminus261, 197]) exercise. 
Second-by-second classification of physical activity across 16090~s of predicted \vo~was able to discern between vigorous, moderate, and light activity with high accuracy (94.1\%).
This system enables quantitative aerobic activity monitoring in non-laboratory settings\added{, when combined with tidal volume calibration,} across a range of exercise intensities using wearable sensors for monitoring exercise prescription adherence and personal fitness.
\end{abstract}

\section{Introduction}

Cardiorespiratory fitness is an established risk factor for cardiovascular disease and all-cause mortality~\cite{aha2016vo2} and is an important determinant for endurance exercise performance \cite{joyner2008endurance}. Cardiorespiratory fitness is conventionally assessed by measuring the rate of oxygen consumption (\vo) and its dynamic response to exercise. Biomarkers such as peak oxygen uptake  (\vopeak) and the rate of adaptation to changes in exercise intensity provide important information about the integrative responses of the pulmonary, cardiovascular, and muscular systems~\cite{wasserman1999book}, which can provide insights into different disease states \cite{eacpraha2012cpx}. Accordingly, \vo~monitoring has become a crucial objective measure for advanced clinical therapies~\cite{mancini1991cpethf}.

\vopeak~is often considered the gold standard metric of cardiorespiratory function. In heart failure, \vopeak~is a strong predictor of 1-year mortality~\cite{mancini1991cpethf}, and is clinically used to select patients for advanced therapies.  In cases where maximal exercise is infeasible, the dynamic response to sub-maximal exercise also provides important indicators of health \cite{BrunnerLaRocca1999,Alexander2003,Schalcher2003,BorghiSilva2012, malhotra2016cpet} and fitness status \cite{Hickson1978,Hagberg1980,Powers1985,Chilibeck1996}. Despite its established importance, monitoring \vo~in non-laboratory settings remains challenging. Direct measurement of \vo~requires a metabolic cart and trained technician, which limits its applicability to laboratory assessment.
Heart rate (HR) has traditionally been used as an inexpensive and non-intrusive proxy to \vo~response to activity and estimate energy expenditure under the assumption that HR varies linearly with \vo~\cite{Swain1998,Strath2000}; however, dynamic \vo~and HR \added{responses} do not always have a direct correspondence with each other, such as following prior exercise~\cite{Bearden2001}.
Thus, ambulatory physiological monitoring using wearable sensors may provide early detection of sub-clinical biomarkers of disease and enable more widespread assessment of cardiorespiratory function~\cite{sana2020wearables}.


Recent advances in wearable technologies and artificial intelligence have led to new developments in non-intrusive cardiorespiratory monitoring. These approaches are generally modeled as regression problems, where a machine learning model learns a transformation function between physiological inputs from wearable sensors and \vo~measured using a gas analyzer system. Earlier work primarily used a combination of HR and activity-related inputs to predict \vo~during various forms of exercise~\cite{beltrame2016treadmill,altini2016vo2}. Sensors embedded in textile fabrics have enabled \vo~prediction in low- to moderate-intensity exercise during activities of daily living~\cite{beltrame2017adl, beltrame2018adl}. Recent studies have leveraged the time-dependent nature of \vo~by modeling the regression as a time-series, or sequential, prediction, where previous physiological states were used to predict \vo~\cite{zignoli2020lstm} and ventilatory threshold~\cite{miura2020vt} during stationary cycling on a cycle ergometer. Although sequential prediction \added{has been shown to} model the temporal nature of oxygen uptake well, \added{further investigations are needed to assess model efficiency and performance across a range of exercise intensities, different days, and demographics}.


In this paper, we propose and evaluate a sequential deep learning model based on temporal convolutional networks (TCN)~\cite{bai2018tcn} for predicting \vo~from physiological inputs derived from smart textiles and a cycle ergometer. The model used causal convolutions to incorporate only past and present physiological response, and the system architecture was designed to provide a tunable effective history length, or receptive fields. We assessed the effect of receptive field and model complexity on prediction accuracy to investigate the temporal relationship between physiological inputs and \vo~response to provide guidance on optimal model design and assessment across a range of different exercise intensities.

\section{Results}

\begin{figure}
    \centering
    \includegraphics[width=0.45\textwidth]{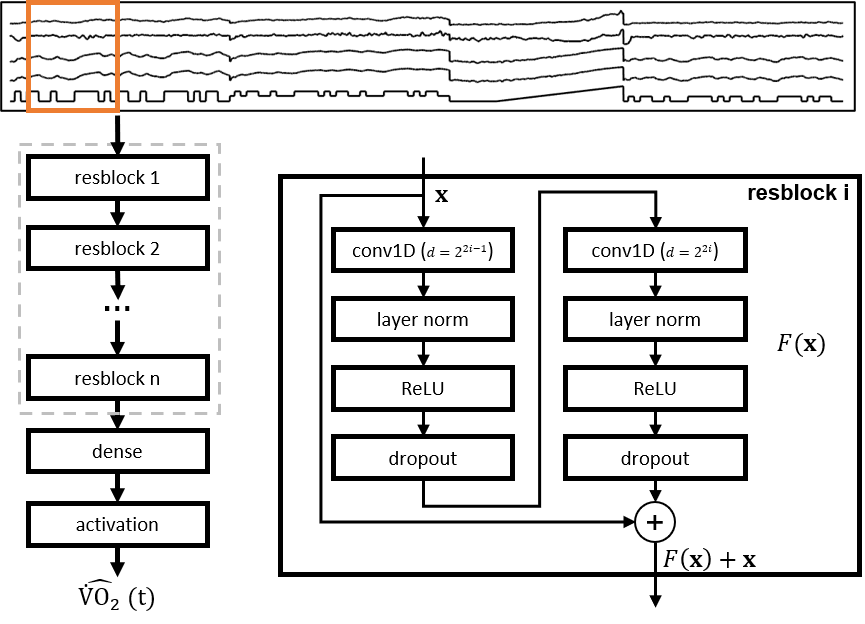}
    \caption{Network architecture. Temporal features (heart rate, heart rate reserve, breathing frequency, minute ventilation, and work rate) are processed through a series of residual blocks with causal convolutions and dilations for feature extraction, followed by a fully connected layer and linear activation to predict \vo~at time $t$.}
    \label{fig:tcn}
\end{figure}

\subsection{Experimental Setup}
Fig.~\ref{fig:tcn} shows the data flow through the TCN network architecture. The four cardiorespiratory biosignals derived from the smart shirt (heart rate (HR), heart rate reserve, breathing frequency, and minute ventilation (\ve))  and the work rate (WR) profile were used as inputs into a chain of residual blocks, followed by a dense layer and linear activation to predict \vo~at each time point.
Results of the TCN networks were compared against a stacked long short-term memory (LSTM) network~\cite{zignoli2020} and random forest (RF)~\cite{beltrame2017rf} prediction models. The stacked LSTM model was trained using the originally proposed features~\cite{zignoli2020lstm}, as well as adding heart rate reserve and \ve~to the feature set, with a sequence length of 140~s approximating 70~breaths at low intensity exercise. The RF model was built using the optimal number of trees according to the validation loss (30~trees; see Supplementary Materials). \vo~data were converted to \si{ml.min^{-1}.kg^{-1}} to compute metabolic equivalent of task (METs) for quantifying physical activity levels (METs=$\dot{\text{V}}\text{O}_2/3.5$~\cite{who2010mets}). METs were classified as light ($<$3.0~METs), moderate (3.0 to 5.9~METs), or vigorous ($\ge$6.0) intensity exercise according to established guidelines~\cite{who2010mets}. 


\subsection{Hyperparameters}
\begin{figure}
    \centering
    \includegraphics[width=0.45\textwidth]{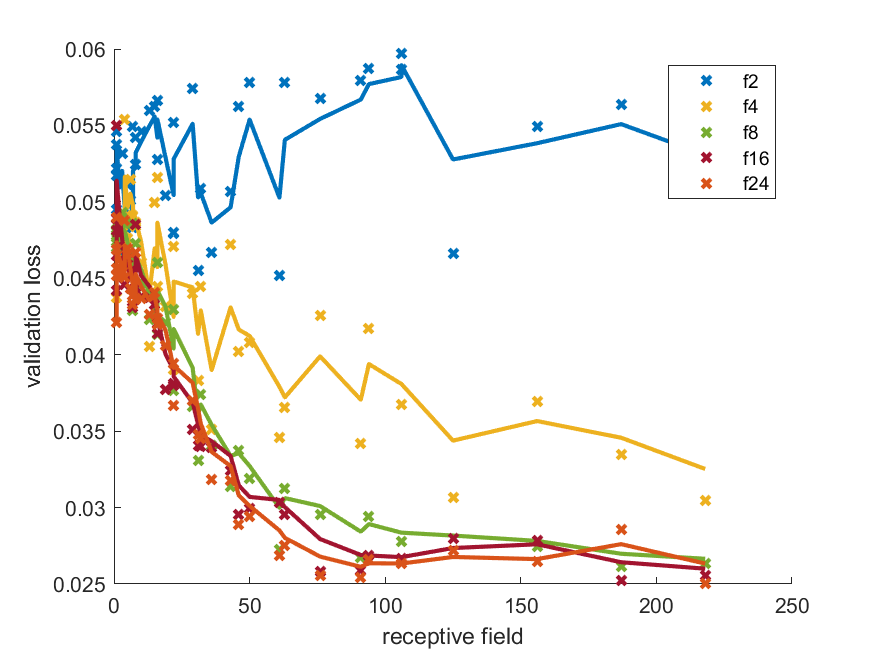}
    \caption{Model validation performance across different receptive fields (determined by kernel size and dilation depth) grouped by the number of filters in the convolutional layers (f\#). Exponential weighted moving average fits were plotted for visualization purposes. The optimal model hyperparameters were 24 filters, 8 kernel sizes, and 5 dilations. Two filters per layer were insufficient for learning the transformation function.}
    \label{fig:results_rf_loss}
\end{figure}

\begin{table}
\centering
\caption{\textsc{Hyperparameter values for investigating the optimal temporal convolutional network configuration}}
\label{tab:hyperparams}
\begin{tabular}{ll}
\hline
\textbf{Hyperparameter} & \textbf{Values} \\ \hline
filters & {2, 4, 8, 16, 24} \\
kernel size & {1, 2, 3, 4, 5, 6, 7, 8} \\
dilation depth & {1, 2, 3, 4, 5} \\ \hline
\end{tabular}
\end{table}

Fig.~\ref{fig:results_rf_loss} shows the effect on validation loss of receptive field (by modifying kernel size $k$ and dilation depth $d$) and number of filters in the causal convolutional layers across a combination of hyperparameter values (see Table~\ref{tab:hyperparams}). Trend lines were generated using exponential weighting moving average ($\alpha=0.5$) for visualization purposes. Validation loss decreased to convergence with larger receptive fields, apart from the two-filter models which were insufficient for learning the prediction function, as evidenced by a flat and highly variable loss curve over all receptive fields. Performance increase was marginal beyond 8~filters. The optimal hyperparameters that produced the smallest hold-out validation loss were 24 filters and a receptive field of 218~s using a kernel size of 8~s and 5~dilations (2 residual blocks). However, a smaller receptive field and/or a smaller set of model parameters may be preferable to a marginal loss increase. Smaller receptive fields allow for reduced ``cold start'' time, and fewer parameters result in decreased computational load. Thus, we assessed the accuracy of models that were within 5\% of the best (minimum) validation loss below.

\subsection{Network Performance}
\begin{figure}
    \centering
    \includegraphics[width=0.43\textwidth]{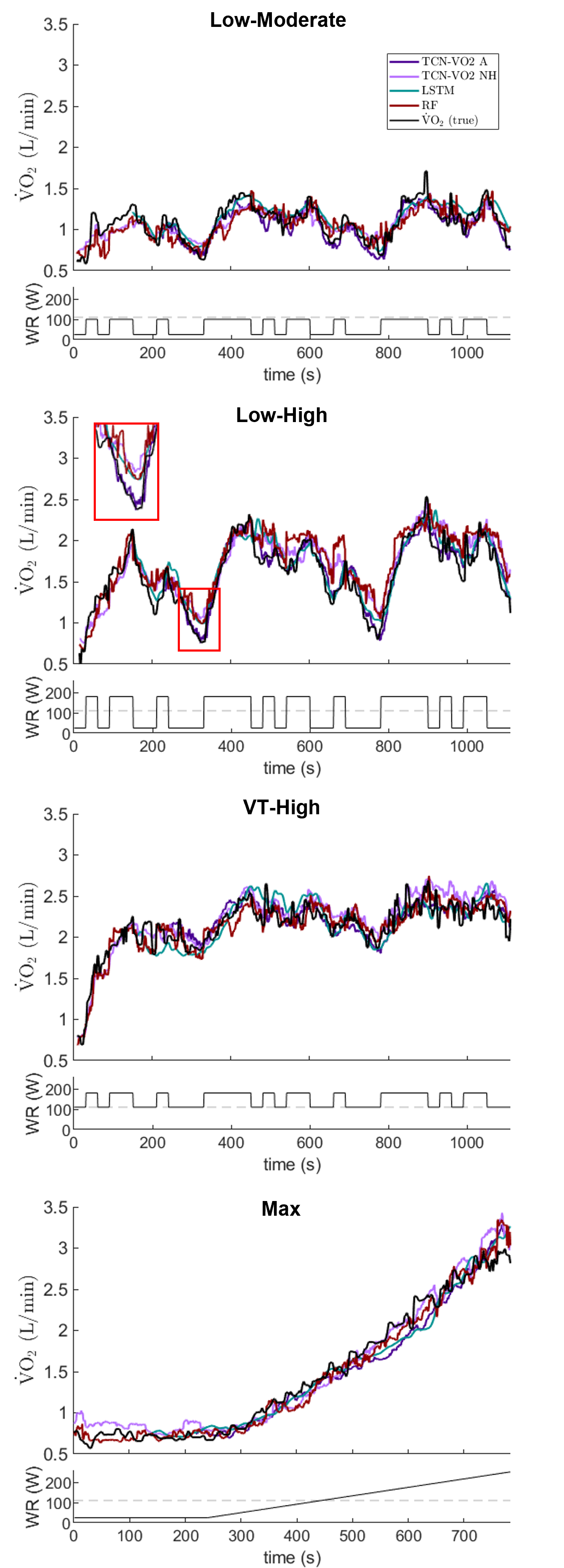}
    \caption{Results for a representative participant for VO2-TCN (proposed) with the lowest validation loss (VO2-TCN A) and 1~s receptive field (VO2-TCN NH), random forest (RF)~\cite{beltrame2017adl}, and stacked long short-term memory network (LSTM)~\cite{zignoli2020lstm} across the four exercise protocols. Dashed gray line represents the participant's ventilatory threshold.}
    \label{fig:results_p01}
\end{figure}

\begin{table*}[]
\centering
\caption{\textsc{Error rates of TCN and comparison models}}
\label{tab:errors}
\begin{tabular}{cccc|c|S[table-format=3(3)]S[table-format=3(3)]S[table-format=3(3)]S[table-format=3(3)]S[table-format=3(3)]S[table-format=3(3), separate-uncertainty]}
\hline
\multicolumn{4}{c|}{\textbf{Method}} & \textbf{\# params} & \multicolumn{6}{c}{\textbf{error (mean $\pm$ SD) [\si{ml.min^{-1}}]}} \\
 &  &  &  &  & {\textbf{L-M}} & {\textbf{L-H}} & {\textbf{VT-H}} & {\textbf{MAX}} & {\textbf{combined}} & {\vopeak} \\ \hline
\multicolumn{4}{c|}{Random forest~\cite{beltrame2017adl}} & 298106 & 20 \pm 134 & 144 \pm 187 & 54 \pm 169 & 15 \pm 111 & 62 \pm 164 & 134 \pm 151 \\
\multicolumn{4}{c|}{Stacked LSTM~\cite{zignoli2020lstm}} & 21589 & 93 \pm 133 & 52 \pm 145 & -58 \pm 148 & -16 \pm 178 & 21 \pm 160 &  -48 \pm 243 \\ 
\multicolumn{4}{c|}{Stacked LSTM+\{HRR,\ve \}} & 21845 & 52 \pm 132 & 7 \pm 163 & -72 \pm 152 & -37 \pm 163 & -10 \pm 159 & -17 \pm 205 \\ 
\hline
 & \textbf{\footnotesize filters} & \textbf{\begin{tabular}[c]{@{}c@{}}{\footnotesize receptive}\\ {\footnotesize field (s)}\end{tabular}} & \textbf{\begin{tabular}[c]{@{}c@{}}{\footnotesize validation}\\ {\footnotesize loss}\end{tabular}} &  &  &  &  &  & &  \\ \hline
\multicolumn{1}{l}{TCN-VO2 A} & 24 & 218 & 0.02506 & 19921 & -23 \pm 116 & 14 \pm 136 & -49 \pm 115 & -32 \pm 117 & -22 \pm 122 & 18 \pm 182 \\
\multicolumn{1}{l}{TCN-VO2 B} & 16 & 187 & 0.02525 & 8081 & -7 \pm 109 & 30 \pm 138 & -25 \pm 115 & -2 \pm 113 & -1 \pm 120 & 62 \pm 162 \\
\multicolumn{1}{l}{TCN-VO2 C} & 16 & 76 & 0.02585 & 5393 & -21 \pm 113 & 16 \pm 159 & -57 \pm 136 & 3 \pm 123 & -17 \pm 137 & 93 \pm 214 \\
\multicolumn{1}{l}{TCN-VO2 D} & 16 & 91 & 0.02586 & 6241 & -24 \pm 115 & 31 \pm 142 & -54 \pm 123 & 9 \pm 117 & -11 \pm 128 & 80 \pm 165 \\
\multicolumn{1}{l}{TCN-VO2 NH} & 24 & 1 & 0.04895 & 361 & 10 \pm 127 & 133 \pm 179 & 69 \pm 129 & 75 \pm 123 & 72 \pm 149 & 223 \pm 196 \\ 
\multicolumn{1}{l}{TCN-VO2 A(HR)} & 24 & 218 & 0.10733 & 19057 & 197 \pm 286 & 133 \pm 373 & 40 \pm 412 & 145 \pm 445 & 127 \pm 379 & -220 \pm 445 \\ \hline
\end{tabular}
\end{table*}

We assessed the performance of models exhibiting validation loss within 5\% of the minimum validation loss, as well as models with no history (TCN-VO2~NH) and using only HR as input (TCN-VO2~A(HR)), and compared the results to existing \vo~prediction methods (Table~\ref{tab:errors}). There were eight models within the 5\% performance threshold, of which we reported four with varied receptive fields and parameter sets. The best performing model according to hold-out validation loss (TCN-VO2~A) had a 218~s receptive field and 19921 parameters. TCN-VO2~B had marginally higher validation loss with 187~s receptive field, indicating minimal performance gain for additional history beyond 187~s. TCN-VO2~C and D had near identical validation loss (3\% increase over model A), but required a receptive field of only 76~s and 91~s respectively, thus providing a reduced cold start period for initial prediction. 

\begin{figure}
    \centering
    \includegraphics[width=0.5\textwidth]{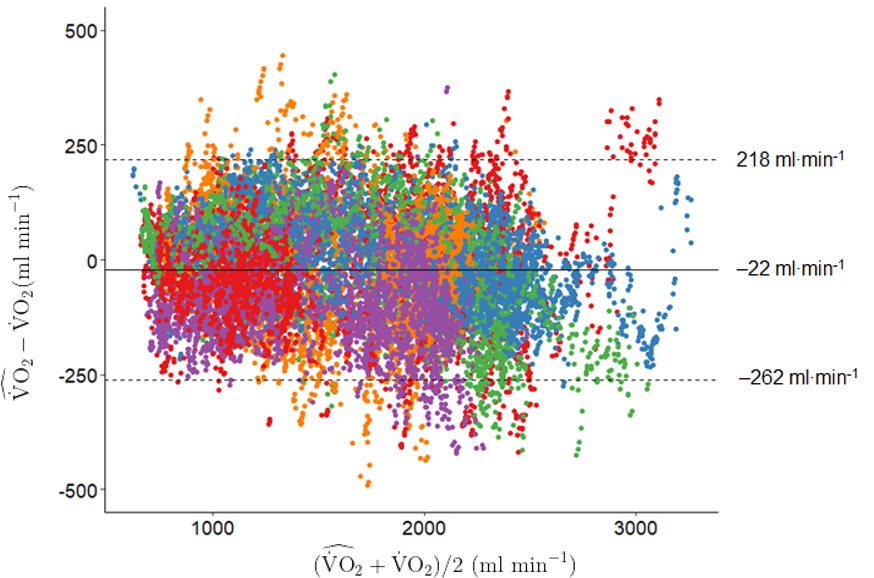}
    \caption{Repeated measures Bland-Altman analysis of the predicted oxygen uptake (\vo) using TCN-VO2 A and directly measured \vo~with all exercise conditions combined. Dotted horizontal lines represent the 95\% limits of agreement and the solid line represents the prediction bias. Each color represents data from a unique participant \added{in the test set}.}
    \label{fig:blandaltman}
\end{figure}

Repeated measures Bland-Altman analysis showed no systematic error between predicted and true \vo~across all TCN-VO2, LSTM and RF models, and no proportional error (Fig.~\ref{fig:blandaltman}). The limits of agreement (LoA) remained relatively constant across TCN-VO2 models during L-M, VT-H, and MAX data, but increased with decreased model complexity in L-H and combined data. Model bias on all combined data was less than 25~\si{ml.min^{-1}} in the best performing TCN-VO2 models and 72~\si{ml.min^{-1}} for 1~s receptive field (model NH), and the equality line fell within the confidence interval of the mean difference. TCN-VO2 models exhibited smaller LoA compared to LSTM in all exercise protocols, indicating a stronger overall fit and smaller error variance. \added{Similarly, mean error of \vopeak~was smallest in TCN-VO2~A (18~\si{ml.min^{-1}}), strongly outperforming models with no history (RF: 134~\si{ml.min^{-1}}, TCN-VO2~NH: 223~\si{ml.min^{-1}}) and HR-only (\textminus220~\si{ml.min^{-1}}), as well as exhibiting smaller LoA than LSTM models.} Furthermore, TCN-VO2 models required substantially fewer network parameters to achieve comparable bias and lower LoA, requiring 1.1$\times$ and 4.1$\times$ fewer parameters in the TCN-VO2~A and TCN-VO2~C, respectively, and 60.5$\times$ fewer in TCN-VO2 NH compared to the stacked LSTM model. TCN-VO2~A(HR) exhibited high error bias and variance across protocols, showing that HR is not a robust estimator of \vo.

TCN-VO2 with no history (``NH''), comprising 1~s receptive field, was compared against RF which performs point-wise predictions without input from previous states. TCN-VO2~NH and RF performed comparably, although the TCN-VO2~NH parameter set contained 825.8$\times$ fewer parameters than RF (quantified as the number of split nodes in the forest). Both TCN-VO2~NH and RF exhibited larger bias in L-H data compared to the other protocols due to overestimating \vo~during low WR transients (see Fig.~\ref{fig:results_p01} inset). \vo~prediction during exercise with relatively small WR transitions was not as severely affected. Top performing TCN models were the only models to correctly predict the lowest \vo~values during the off-transients, making them good candidates for assessing kinetics.

\subsection{Quantifying Physical Activity Levels}
\begin{figure*}
    \centering
    \includegraphics[width=\textwidth]{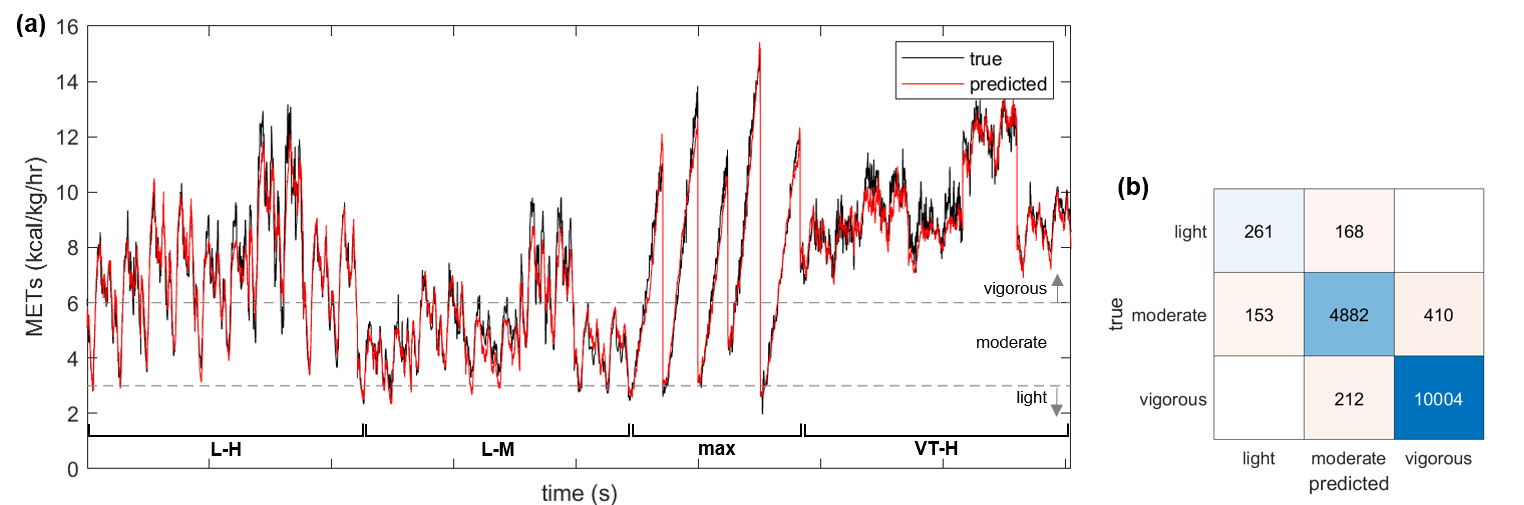}
    \caption{(a) Prediction of metabolic equivalent of task (METs) over all test data (4~protocols per participant across 5~participant \added{test set}) concatenated into one plot and visually grouped by protocol for quantifying physical activity levels according to global guidelines. (b) Confusion matrix of physical activity classification (light: $<$3.0; moderate: 3.0 to 5.9; vigorous: $\ge$6.0)}
    \label{fig:results_confmat}
\end{figure*}

Predicted \vo~were used in accordance with globally established guidelines on exercise prescription for health~\cite{who2010mets} to classify second-by-second activity levels. These guidelines are often used in practice, alongside other biomarkers, for optimizing cardiovascular health during rehabilitation exercise prescription. Fig.~\ref{fig:results_confmat} shows the METs data derived from the true (measured) and predicted \vo~data, as well as the confusion matrix. All test data (4 protocols/participant across the 5-participant test set) were classified on a second-by-second basis and visually concatenated into a single plot spanning a total of 16090~s of predicted metabolic activity across 20~individual exercise trials. Overall, 15147~s (94.1\%) of the 1~Hz data were correctly classified into appropriate physical activity categories. The data spanned across light (2.7\%), moderate (33.8\%), and vigorous (63.5\%) intensity exercise. During moderate-intensity exercise (5445~s), 4882~s (89.7\%) were correctly classified, while 153~s (2.8\%) and 410~s (7.5\%) were classified as light and vigorous activity.
During vigorous-intensity exercise (10216~s), 10004~s (97.9\%) were correctly classified, while 212~s (2.1\%) were classified as moderate activity.
Less than 3\% of the data were in the ``light" category (429~s), and of those data, 261~s (60.8\%) were correctly classified and 168~s (39.2\%) were classified as moderate activity. 90\% of these data were within 0.5~METs of the threshold.

\section{Discussion}
This work has shown that the complex \added{dynamic} \vo~\added{response} to \added{changes in} exercise intensity can be accurately predicted using sequential deep learning models across a range of low-, moderate-, and \added{high}-intensity exercise, as well as maximal aerobic exercise. The baseline model TCN-VO2~NH (with 1~s receptive field) and point-wise RF regression~\cite{beltrame2017adl} were both blind to previous states, using only the current time point for prediction. In both cases, we observed a large and similar bias in the L-H data, albeit being two fundamentally different model types. The source of error was largely due to overestimation during recovery in the off-transients. The overestimation is likely a result of the point-wise predictor's blindness to the previous state, as the models were unable to adequately learn the relationships between model inputs and outputs during on- and off-transients. This is an important consideration, as differences in HR \added{dynamics} have been observed between on- and off-transients, with the slower off-transients being amplified following \added{high}-intensity exercise \cite{Bearden2001,Linnarsson1974}. HR kinetics are also reported to be slower than \vo~kinetics during recovery \cite{Bearden2001}. Similarly, the ventilation response has been observed to be slower during recovery compared to exercise onset~\cite{Hughson1995}, with the rate of change in ventilation being markedly slowed when recovering from higher intensity exercise \cite{Linnarsson1974}. Furthermore, the ventilatory response is slower than the \vo~response \cite{Bell1999,HughsonMorrissey1982}. Accordingly, without knowing the previous history of the system, these point-wise models are naive to status of the system leading to erroneous \vo~predictions.
Thus, it appears that the temporal models were able to more accurately learn the relationships between HR and ventilation, and \vo~during exercise onset as well as recovery.

The best performing models, when ranked by validation loss, were consistently those with receptive fields of 218~s, 187~s, 76~s, and 91~s. We observed marginal difference between the test errors of these models, although there were some differences in predicting \vo~minima at the end of off-transients, which did not significantly affect the global loss function. Considering that the standard \vo~time constant is typically 20--30~s in healthy populations \cite{poole2011kinetics}, these receptive fields range from approximately 3-11 time constants, which suggests that the best models tend to use a receptive field that includes most, if not all, of the transient phase for a sustained step change in WR to achieve a new constant \vo. Receptive fields of 76~s and 91~s appear consistent with the protocol's longest off-transient (90~s). The longest on-transient is 120~s. Thus, further investigations are needed to determine the effects of receptive field during different exercise protocols.

Both classes of sequential deep learning models (TCN and LSTM) exhibited strong predictive power. TCN architectures have become popular alternatives to recurrent neural network instances largely due to their data parallelism, flexible receptive field size, and stable gradients~\cite{bai2018tcn}. In this work, top performing TCN models were much smaller than the stacked LSTM models, yielding more computationally efficient prediction networks. 

The \vo~predictions were converted to METs, which is an established metric for quantifying physical activity levels~\cite{who2010mets}. Quantifying activity levels is helpful for exercise prescription in cardiovascular disease management~\cite{wahid2016mets}. However, traditional patient recall may be affected by recall and/or social desirability bias~\cite{althubaiti2016bias}, providing uncertainty in an important biomarker for cardiovascular health. This work showed strong accuracy across a range of METs categories, which provides supporting evidence for quantitative and objective at-home activity monitoring using wearable sensors. The errors in the light activity category were mainly during category transition, not sustained activity. In these data, the majority of light data was close to threshold. It would be expected that incorporating data from resting of very low-intensity exercise would lead to a more representative error profile.

The primary limitations impacting the widespread generalizability of these results stem from the dataset's constrained demographic (young healthy adults) and structured exercise protocol. 
We selected the three-stage PRBS protocol because it challenged the dynamic response of aerobic metabolism across a wide range of intensities that could occur in real-life situations. PRBS exercise provides an opportunity to directly quantify an index of physical fitness~\cite{Hedge2020}; however, PRBS cycling exercise is a controlled laboratory protocol. Accordingly, our optimization results may not be directly applicable to unstructured exercise or different exercise protocols (i.e., constant load exercise). Further investigations in different exercise situations with a more diverse participant sample, including cardiovascular-related diseases, are needed to assess its generalizability to different kinds of physical activity and out-of-sample populations. \added{Additionally, the exercise in this study was performed in a temperature controlled environment. Further investigations are needed to evaluate model performance in specialized environments that may alter cardiovascular response to exercise (e.g., hypoxia, heat stress, etc.)}.




To conclude, using causal convolutions in a temporal deep learning model, the effect of receptive field (i.e., effective history) on \vo~prediction was assessed by performing a grid search across hyperparameter values. The best performing models, according to validation loss, comprised receptive fields of 218~s, 187~s, 97~s, and 76~s. Results showed low prediction error across a wide range of exercise intensities with drastically reduced parameter sets compared to existing methods. Using HR as the only input feature into the same model architecture yielded substantially larger errors, reinforcing that HR alone is insufficient for predicting \vo, thus necessitating more complex approaches. Using the temporal prediction outputs, physical activity levels were quantified to provide a breakdown of time spent in light, moderate, and vigorous activity according to global health guidelines. These results suggest that cardiorespiratory function may be assessed in non-laboratory settings\added{, when combined with tidal volume calibration,} across a wide range of activity levels using wearable sensors and smart textiles.

\section{Methods}
\subsection{Data Collection and Preprocessing}
Twenty-two young healthy adults (13 males, 9 females; age: 26 $\pm$ 5 \si{yr}; height: 1.71 $\pm$ 0.08 \si{m}; mass: 70 $\pm$ 11 \si{kg}; \vopeak: 42 $\pm$ 6 \si{ml.min^{-1}.kg^{-1}}) with no known musculoskeletal, respiratory, cardiovascular, or metabolic conditions volunteered to participate in the study. The study was approved by a University of Waterloo Research Ethics committee (ORE~\#32164) and conducted in accordance with the Declaration of Helsinki. All participants signed an informed consent before participating.

Participants visited the laboratory on four separate occasions to perform a ramp-incremental exercise test, and three different pseudorandom binary sequence (PRBS) exercise tests~\cite{Hughson1990prbs}. Each exercise session was separated by at least 48~\si{h}, and participants were instructed to arrive for testing at least two hours postprandial, and abstain from alcohol, caffeine and vigorous exercise in the 24~hours preceding each test. All exercise tests were performed in an environmentally controlled laboratory on an electronically braked cycle ergometer (Lode Excalibur Sport, Lode B.V., Groningen, Netherlands). Participants were instructed to maintain cadence at 60 revolutions per minute for all exercise tests.

\begin{figure*}
    \centering
    \includegraphics[width=\textwidth]{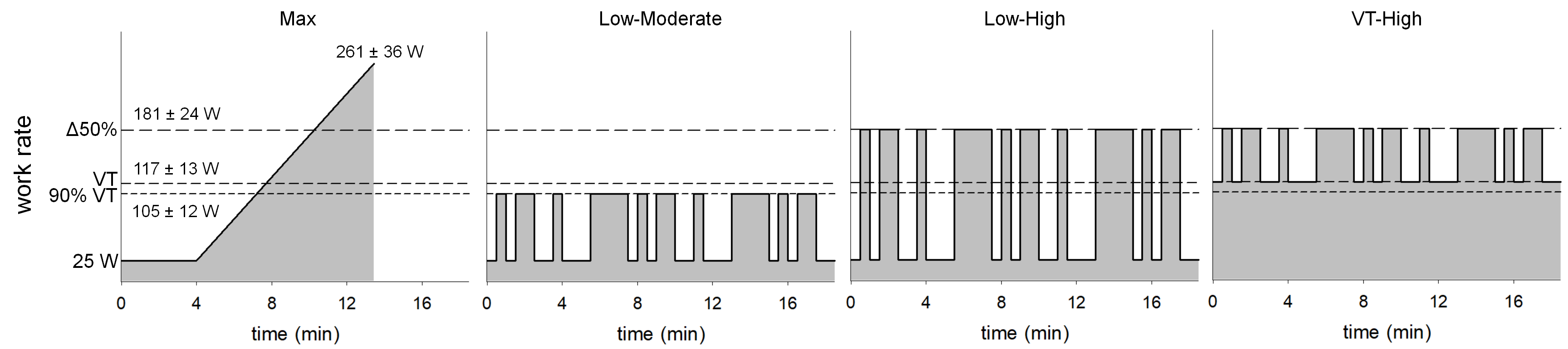}
    \caption{Schematic showing the incremental ramp and three pseudorandom binary sequence (PRBS) cycling tests. PRBS protocols were designed using work rates at 90\% ventilatory threshold (VT), VT, and the midpoint between VT and \vopeak~($\Delta$50\%) using participant-specific VT and \vopeak~determined from the ramp test.}
    \label{fig:WRprofile}
\end{figure*}

On the first visit, 5~\si{min} of seated resting data were collected to determine each participant’s resting HR. After the resting period, participants performed a ramp-incremental exercise test to exhaustion (25~\si{W} baseline for 4~\si{min} followed by a 25~\si{W.min^{-1}} ramp) to determine each participant’s ventilatory threshold (VT) \cite{Beaver1986}, \vopeak, and the work rates (WR) for the PRBS exercise tests (Fig.~\ref{fig:WRprofile}). The test was terminated when the cadence dropped below 55 revolutions per minute despite strong verbal encouragement. \vopeak~was defined as the highest \vo~computed from a 20~\si{s} moving average during the exercise test. \vo~at VT was estimated by visual inspection using standard ventilatory and gas exchange indices, and their ratios, as previously described \cite{Beaver1986}. WRs at 90\% VT, VT, and the midpoint between VT and \vopeak~(referred to as $\Delta$50\%) were estimated by left-shifting the \vo~response by each individual’s mean response time to align the \vo~and WR profiles. \added{Mean response time was determined by fitting a double-linear model to the ramp-incremental data, and finding the point of intersection between the forward extrapolation of the average \vo~during the 25~W baseline cycling in the 2~min prior to ramp onset, and the backwards extrapolation of the linear portion of the ramp \vo~response below VT~\cite{Keir2018}.}

In visits 2-4, participants performed one of three different PRBS exercise tests in a randomized order. WRs systematically alternated in the three PRBS exercise test between 25~\si{W} and 90\% of the VT (low-to-moderate; L-M), 25~\si{W} and $\Delta$50\% (low-to-\added{high}; L-H), or VT and $\Delta$50\% (VT-to-\added{high}; VT-H). The time series for the changes in WR for PRBS protocols were generated by a digital shift register with an adder module feedback \cite{BeltrameMNG,BeltrameLinear,Bennett1981,Hughson1990prbs}. This process pseudo-randomized the changes in WR and ensured that there would be sufficient \vo~signal amplitude while performing non-constant load exercise. A single PRBS was composed of 15 units, each of 30~\si{s} in duration, totaling 7.5~\si{min}. Each complete PRBS testing session consisted of a 3.5~\si{min} warm-up (the last 3.5~\si{min} of the 7.5~\si{min} PRBS), and then two full repetitions of the PRBS for a total of 18.5~\si{min} of continuous cycling per session.

A portable metabolic system (MetaMax 3B-R2, CORTEX Biophysik, Leipzig, Germany) was used to measure gas exchange during all exercise tests. Participants breathed through a mask (7450 SeriesV2 Mask, Hans Rudolph Inc., Shawnee, KS, USA), and inspired and expired flow were measured using a bi-directional turbine. The turbine was calibrated before each testing session using a 3\si{L} syringe. Oxygen and carbon dioxide gas concentrations were continuously sampled at the mouth and were analyzed using a chemical fuel cell and nondispersive infrared sensor, respectively. Precision-analyzed gas mixtures were used to calibrate the oxygen and carbon dioxide gas concentrations. \vo~and carbon dioxide output were calculated using standard breath-by-breath algorithms~\cite{wasserman1999book}. \vo~data were filtered using a sliding 5-breath median filter to correct spurious outlier breaths.

Participants wore a HR monitor (Polar H7, Polar Electro Oy, Kempele, Finland) that wirelessly communicated with the portable metabolic system, such that HR data were logged synchronously with the gas exchange data. Participants were also fitted with a wearable integrated sensors shirt that was sized to each participant based on manufacturer guidelines (Hexoskin, Carre Technologies, Montreal, Canada). The shirt contained a textile electrocardiogram to measure HR, and thoracic and abdominal respiration bands to obtain estimates of breathing frequency and minute ventilation (\ve) via respiratory inductance plethysmography. \added{Estimates of \ve~provided by the smart shirt were calibrated by linear regression to the known \ve~measured throughout each protocol with the bi-directional turbine.} The estimates of HR, \ve, and breathing frequency from the smart shirt have been previously validated~\cite{Villar2015}. Data recorded by the metabolic system and smart shirt were time-aligned by cross-correlating the two different HR signals. After processing, all data were interpolated to 1~Hz to ensure signal synchronization and a constant sampling rate. The final set of physiological features were: work rate (\si{W}), \ve~(\si{L.min^{-1}}), breathing frequency (\si{breath.min^{-1}}), HR (\si{bpm}), and heart rate reserve (\%). \added{Heart rate reserve at time $t$ was calculated as $(\text{HR}_\text{t}-\text{HR}_\text{rest})/(\text{HR}_\text{max}-\text{HR}_\text{rest})$, where $\text{HR}_\text{rest}$ and $\text{HR}_\text{max}$ were determined during 5~min rested seated baseline and the peak HR during ramp-incremental test, respectively.}


\subsection{Network Architecture}

\vo~kinetics have a systematic, albeit complex, temporal response to exercise~\cite{whipp2005}. 
Exponential models for quantifying \vo~kinetics during controlled exercise protocols have demonstrated that the intensity of the exercise being performed strongly influences the dynamic \vo~response~\cite{Koppo2004,McNarry2012,Hedge2020}. Thus, as an alternative to conventional convolutional neural networks that use a symmetric kernel about the current time (or space) location, we developed a sequential convolutional model using causal convolutions~\cite{bai2018tcn}. Specifically, given a sequence of time series inputs $\vec{x}_1, \vec{x}_2, \ldots, \vec{x}_T \in \mathbb{R}^n$ extracted from wearable sensors and a cycle ergometer, the goal was to predict the \vo~at time $t$, denoted $\dot{\text{V}}\text{O}_{2,t}$. Specifically, given a prediction model $M:\mathcal{X}^T \rightarrow \mathcal{Y}$, $\dot{\text{V}}\text{O}_{2,t}$ is predicted using historical temporal features $\vec{x}_i \in \mathbb{R}^n$ only up to the current time point:
\begin{equation}
    \widehat{\dot{\text{V}}\text{O}}_{2,t} = M(\vec{x}_t, \vec{x}_{t-1}, \ldots, \vec{x}_{t-w})
\end{equation}
where $w$ is the effective history. 

Our model incorporated a temporal feature extraction network and a regression network. The feature extraction network was implemented as a temporal convolutional network (TCN) with a tunable receptive field through multiple sequential layers with kernel dilation for multi-scale aggregation of the input data~\cite{bai2018tcn,yu2016dilated,oord2016wavenet}. The TCN was implemented as a sequence of stacked residual blocks comprised of repetitions of dilated causal 1D convolution, layer normalization, rectified linear activation, and dropout (see Fig.~\ref{fig:tcn}). The input into the residual block $\vec{x}$ was added to the residual function $F(\vec{x})$ through identity mapping (or 1x1 convolution in the first residual block when the number of channels did not match the residual function shape) to encourage learning of the residual modifications of the input data, which has been shown to improve the performance of deep networks~\cite{he2016resnet}. Each residual block is composed of successive pairs of dilations, rather than applying the same dilation twice as in the original TCN description~\cite{bai2018tcn}, for better control of the receptive field. In models with an odd number of dilations, the first three were grouped into a single residual block, inspired by ResNet, which learns a function over two or three layers~\cite{he2016resnet}. Thus, the (causal) receptive field was determined by the kernel size $k$ and the number of exponential dilations $N$:
\begin{equation}
    RF=1+(k-1)(2^N-1)
    \label{eq:rf}
\end{equation}

A splice function was used to extract the features at the last known time point $t$ for input into the regression network. This network was defined simply as a fully connected dense layer and linear activation to predict $\dot{\text{V}}\text{O}_{2,t}$.


\subsection{Training and Hyperparameter Optimization}
The data across all 22~participants were split into train (40 protocols), test (20 protocols), and validation (19 protocols) datasets. Each participant's whole data were only comprised within one of train, test, or validation, with no participant data spread to encourage generalizability. Data of specific sequence lengths (according to the receptive field) were extracted from whole exercise protocols through a sliding window method, where the first sequence of sequence length $T$ was extracted from the onset of exercise to time $T$, and subsequent sequences were extracted until the end of the exercise was reached. This yielded a dataset of size $\mathbb{R}^{N\times T \times F}$, where $N$ is the number of sequences, $T$ is the sequence length, and $F$ is the number of feature signals. The sequence length was determined by the network's receptive field. Each feature, except WR, was standardized to zero mean and unit variance according to the training data statistics. WR was normalized to [0, 1] due to its non-normal distribution. 

The hyperparameters of importance are the number of convolutional filters and the receptive field, which is defined by kernel size, dilation rate and network depth (Eq.~(\ref{eq:rf})). This has practical implication in that having a very long receptive field results in a delayed initial prediction (or ``cold start'') of that length of time. Hyperparameter search (number of filters, kernel size, dilation depth) was performed on the hold-out validation set. Table~\ref{tab:hyperparams} lists the hyperparameter values that were searched. A grid search was performed on a distributed computing cluster using 108~cores and 36~NVIDIA T4 Turing GPUs across 9~compute nodes.

The network was trained using the Adam optimizer, 0.2 dropout rate, 32 minibatch size, and a learning rate of 0.0005 over 100 epochs. For each hyperparameter combination, the epoch with the lowest validation loss was saved. The network hyperparameters producing the lowest mean squared error in the validation set was chosen as the final network architecture.

\subsection{Data Analysis}
Signals were analyzed in MATLAB (2020b, MathWorks, Portola Valley, CA, USA). Statistical analyses were conducted in R (version 3.5.1). The agreement between the predicted and directly measured \vo~were assessed using repeated measures Bland-Altman analysis, which accounts for the within-participant variance of the repeated measures data \cite{BlandAltmanRM}. \added{\vopeak~agreement was assessed using standard Bland-Altman.}

\section{Acknowledgment}
This work was supported by the Natural Sciences and Engineering Research Council of Canada (RGPIN-6473, PDF-503038-2017), and the Canadian Institutes of Health Research Banting and Best Canada Graduate Scholarship (201911FBD-434513-72081). This work was made possible by the facilities of the Shared Hierarchical Academic Research Computing Network (SHARCNET) and Compute/Calcul Canada.

\section{Author Contributions}
R.A., E.T.H. and R.L.H. designed the study. E.T.H. collected the data. R.A. developed the machine learning code and related analyses. E.T.H. conducted the statistical analyses. R.A. and E.T.H. independently validated the results. R.A. and E.T.H. wrote the first draft of the manuscript. All authors revised and approved the final manuscript.

\section{Competing Interests}
All authors declare no competing interests.

\bibliographystyle{IEEEtran}

\end{document}